\documentclass[10pt,twocolumn,letterpaper]{article}

\usepackage{iccv}
\usepackage{times}
\usepackage{epsfig}
\usepackage{graphicx}
\usepackage{amsmath}
\usepackage{amssymb}
\usepackage{multirow}
\usepackage{enumitem}

\usepackage[pagebackref=true,breaklinks=true,letterpaper=true,colorlinks,bookmarks=false]{hyperref}

\iccvfinalcopy 

\def\iccvPaperID{6781} 

\ificcvfinal\fi

\begin{document}

\title{DESOBAv2: Towards Large-scale Real-world Dataset for Shadow Generation}

\author{Qingyang Liu, Jianting Wang, Li Niu\thanks{Corresponding author.} \\
Department of Computer Science and Engineering, MoE Key Lab of Artificial Intelligence, \\
Shanghai Jiao Tong University\\
{\tt \small \{narumimaria,glory1229,ustcnewly\}@sjtu.edu.cn}
}

\maketitle
\ificcvfinal\thispagestyle{empty}\fi


\begin{abstract}
Image composition refers to inserting a foreground object into a background image to obtain a composite image. In this work, we focus on generating plausible shadow for the inserted foreground object to make the composite image more realistic. To supplement the existing small-scale dataset DESOBA, we create a large-scale dataset called DESOBAv2 by using object-shadow detection and inpainting techniques. Specifically, we collect a large number of outdoor scene images with object-shadow pairs. Then, we use pretrained inpainting model to inpaint the shadow region, resulting in the deshadowed images. Based on real images and deshadowed images, we can construct pairs of synthetic composite images and ground-truth target images. Dataset is available at \href{https://github.com/bcmi/Object-Shadow-Generation-Dataset-DESOBAv2}{https://github.com/bcmi/Object-Shadow-Generation-Dataset-DESOBAv2}.
\end{abstract}


\section{Introduction} \label{sec:intro}
Image composition refers to cutting out a foreground object and pasting it on another background image to acquire a composite image, which could benefit plenty of applications in art, movie, and daily photography \cite{ic_useful1,ic_useful2,ic_useful3}. However, the quality of composite images could be significantly compromised by the inconsistency between foreground and background, including appearance, geometric, and semantic inconsistencies. 
With the increasing popularity of deep learning, many deep learning models \cite{ic_full1,ic_full2,ic_full3} have endeavored to tackle different types of inconsistencies in composite images, but only a few works \cite{sgrnet} focused on the shadow inconsistency, which is a crucial aspect of appearance inconsistency. In this work, we aim to cope with the shadow inconsistency, \emph{i.e.}, generating plausible shadow for the foreground object to make the composite image more realistic.

\begin{figure}[t]
\centering
\includegraphics[width=1.0\linewidth]{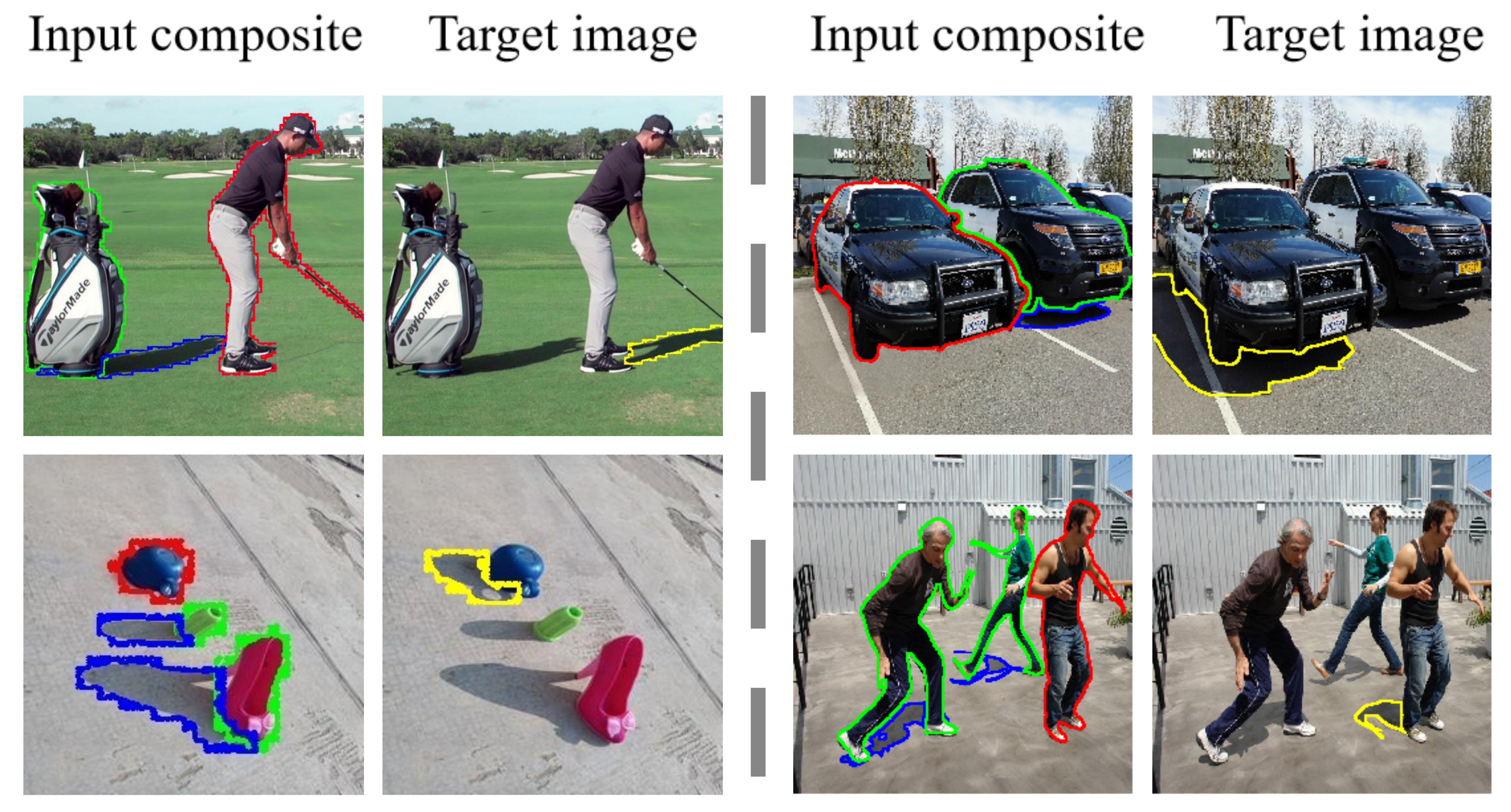}
\caption{The example data for shadow generation task. The left two examples are from our DESOBAv2 dataset and the right two examples are from DESOBA dataset \cite{sgrnet}. In each example, the background object (\emph{resp.}, shadow) mask is outlined in {\color{green}green} (\emph{resp.}, {\color{blue}blue}) and the foreground object (\emph{resp.}, shadow) mask is outlined in {\color{red}red} (\emph{resp.}, {\color{yellow}yellow}). }
\label{fig:example}
\end{figure}

For the shadow generation task, the used data are exhibited in Figure \ref{fig:example}. Given a composite image $I_c$ without foreground shadow, the foreground object mask $M_{fo}$, and the masks of background object-shadow pairs $\{M_{bo},M_{bs}\}$, our goal is generating $\hat{I}_g$ with foreground shadow to make a realistic composite image. To solve this problem, SGRNet~\cite{sgrnet} released the first synthetic dataset DESOBA for real-world scenes. The data in DESOBA can be summarized as tuples of the form $\{I_c,M_{fo},M_{fs},M_{bo},M_{bs},I_g\}$, which include the ground-truth target image $I_g$ and foreground shadow mask $M_{fs}$. The way to obtain pairs of composite images and ground-truth images in DESOBA is as follows. First, they manually remove the shadows in real images to get shadow-free images. Then, they replace one foreground shadow region in a real image (ground-truth image) with the counterpart in its shadow-free version, yielding a synthetic composite image without foreground shadow.  \textbf{Due to the high cost of manual shadow removal, the scale of DESOBA dataset is very limited} (\emph{e.g.}, 1012 ground-truth images and 3623 tuples). Nevertheless, deep learning models require abundant training data. 

To supplement DESOBA dataset, we create a large-scale dataset DESOBAv2 using object-shadow detection and inpainting techniques. Specifically, we first collect a large number of real-world images in outdoor scenes with natural lighting. Then we use pretrained object-shadow detection model \cite{Wang_2022_TPAMI} to predict object and shadow masks for object-shadow pairs. Next, we use pretrained inpainting model \cite{Rombach_2022_CVPR} to inpaint the detected shadow regions to get deshadowed images. Finally, based on real images and deshadowed images, we construct pairs of synthetic composite images and ground-truth target images. The dataset construction details and dataset statistics can be found in Section \ref{sec:DESOBAv2} and Section \ref{sec:statistics}, respectively.

\section{Dataset Construction} \label{sec:DESOBAv2}
In this section, we will introduce the details of constructing our DESOBAv2 dataset. 

\subsection{Shadow Image Collection}
We harvest an extensive collection of real-world outdoor images with natural lighting across various scenes from the Internet. We manually filter the collected images according to the following rules: 1) The images should be realistic images genuinely captured by camera without been manually manipulated. 2) Each image should contain at least one object-shadow pair. 3) The objects casting shadows should be almost complete in the image, but their shadows can be incomplete. The shadows from external objects outside the image should only occupy a small part of the whole image. 4) There should be no large overlaps between the shadows of different objects. 5) There should be no significant occlusion between the objects with shadows. 6) The objects casting shadows should be on the ground instead of off the ground (\emph{e.g.}, a jumping person). 7) All shadows should be cast on the ground and not on vertical surfaces like walls or windows. 8) The shooting angle should be top-down or straight-on.

\subsection{Shadow Removal}
Given a real image, we use pretrained object-shadow detection model \cite{Wang_2022_TPAMI} to predict object and shadow masks for object-shadow pairs. \textbf{We refer to one detected object-shadow pair as one detected instance.} We eliminate the images without any detected instance. 

Subsequently, we attempt to erase all the detected shadows. We have tried some state-of-the-art shadow removal models like \cite{shadowformer}, but the performance in the wild is far from satisfactory due to poor generalization ability. Due to the recent advance of image inpainting models trained on large-scale datasets, we resort to image inpainting to remove the shadows. Although image inpainting cannot preserve the background information, we observe that the background textures in the shadow region are usually very simple, and the inpainted result has similar textures with the original background. Thus, \textbf{we roughly treat the inpainted results as deshadowed results.}

We obtain the union of all detected shadow masks as the inpainting mask and apply the pretrained inpainting model \cite{Rombach_2022_CVPR} to get a deshadowed image $I_d$. 
In practice, we observe that the inpainting model is prone to generate low-quality shadow in the inpainted region in some cases. 
To prevent the inpainting model from generating shadows in the inpainted region, we adopt some tricks like dilating the inpainting mask and flipping images vertically.  However, the inpainting model may still generate shadows or noticeable artifacts in the inpainted region. 

After inpainting, we manually filter the object-shadow pairs according to the following rules: 1) We remove the object-shadow pairs with low-quality object masks or shadow masks. 2) We remove those object-shadow pairs with generated shadows or noticeable artifacts in the shadow region. After manual filtering, \textbf{we refer to the remaining object-shadow pairs as valid instances}.  

\begin{figure*}[p]
\centering
\includegraphics[width=0.9\linewidth]{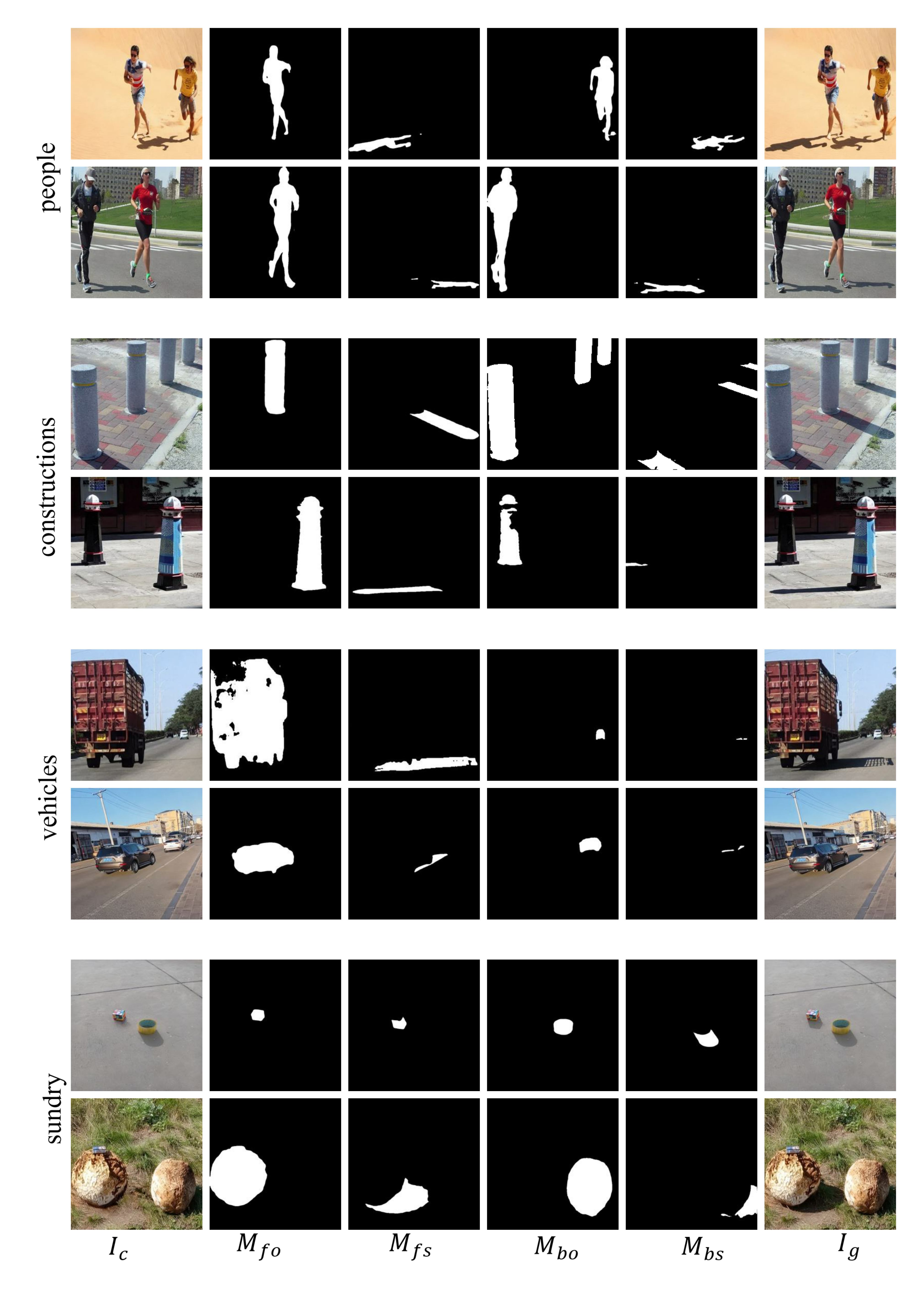}
\caption{Some examples of our DESOBAv2 dataset.}
\label{fig:moreRENOS}
\end{figure*}

\subsection{Composite Image Synthesis}

Given a pair of real image $I_r$ and deshadowed image $I_d$, we randomly select a foreground object from valid instances and synthesize the composite image. One approach is replacing the shadow region of this foreground object in $I_r$ with the counterpart in $I_d$ to erase the shadow of foreground object. However, this approach may leave traces along the shadow boundary so that the model could find a shortcut to generate the shadow. 
Another approach is replacing the shadow regions of the other objects in $I_d$ with the counterparts in $I_r$ to synthesize a composite image $I_c$, in which only the selected foreground object does not have shadow and all the other objects have shadows. We adopt the second approach. 

After inpainting, the pixel values in the background may be slightly changed, that is, the background of  $I_c$ could be slightly different from that of $I_r$. To ensure consistent background, we obtain the ground-truth target image $I_g$  by replacing the shadow regions of all objects in $I_d$ with the counterparts in $I_r$. Then, $I_c$ and $I_g$ form a pair of input composite image and ground-truth target image.

Given an image with $K$ detected instances, we use $M_{o,k}$ (\emph{resp.}, $M_{s,k}$) to denote the object (\emph{resp.}, shadow) mask of the $k$-th object. When choosing the $k$-th object as foreground object,  $M_{o,k}$ (\emph{resp.}, $M_{s,k}$) is 
the foreground object (\emph{resp.}, shadow) mask $M_{fo}$ (\emph{resp.}, $M_{fs}$). We can merge $\{M_{o,1}, \ldots, M_{o,k-1}, M_{o,k+1}, \ldots, M_{o,K}\}$ as the background object mask $M_{bo}$. Similarly,  we can merge $\{M_{s,1}, \ldots, M_{s,k-1}, M_{s,k+1}, \ldots, M_{s,K}\}$ as the background shadow mask $M_{bs}$. Up to now, we obtain a tuple in the form of $\{I_c,M_{fo},M_{fs},M_{bo},M_{bs},I_g\}$, which is consistent with the tuple format in DESOBA dataset. 


We provide some examples from our DESOBAv2 dataset in Figure \ref{fig:moreRENOS}. For each super category of foreground objects (people, constructions, vehicles, sundry), we show two tuples in the form of $\{I_c,M_{fo},M_{fs},M_{bo},M_{bs},I_g\}$.

\section{Dataset Statistics} \label{sec:statistics}
After all the filtering process in Section~\ref{sec:DESOBAv2}, we eventually have $21,575$ images with $28,573$ valid instances, in which each image has at least one valid instance. 

\begin{figure}[t]
\centering
\includegraphics[width=1.0\linewidth]{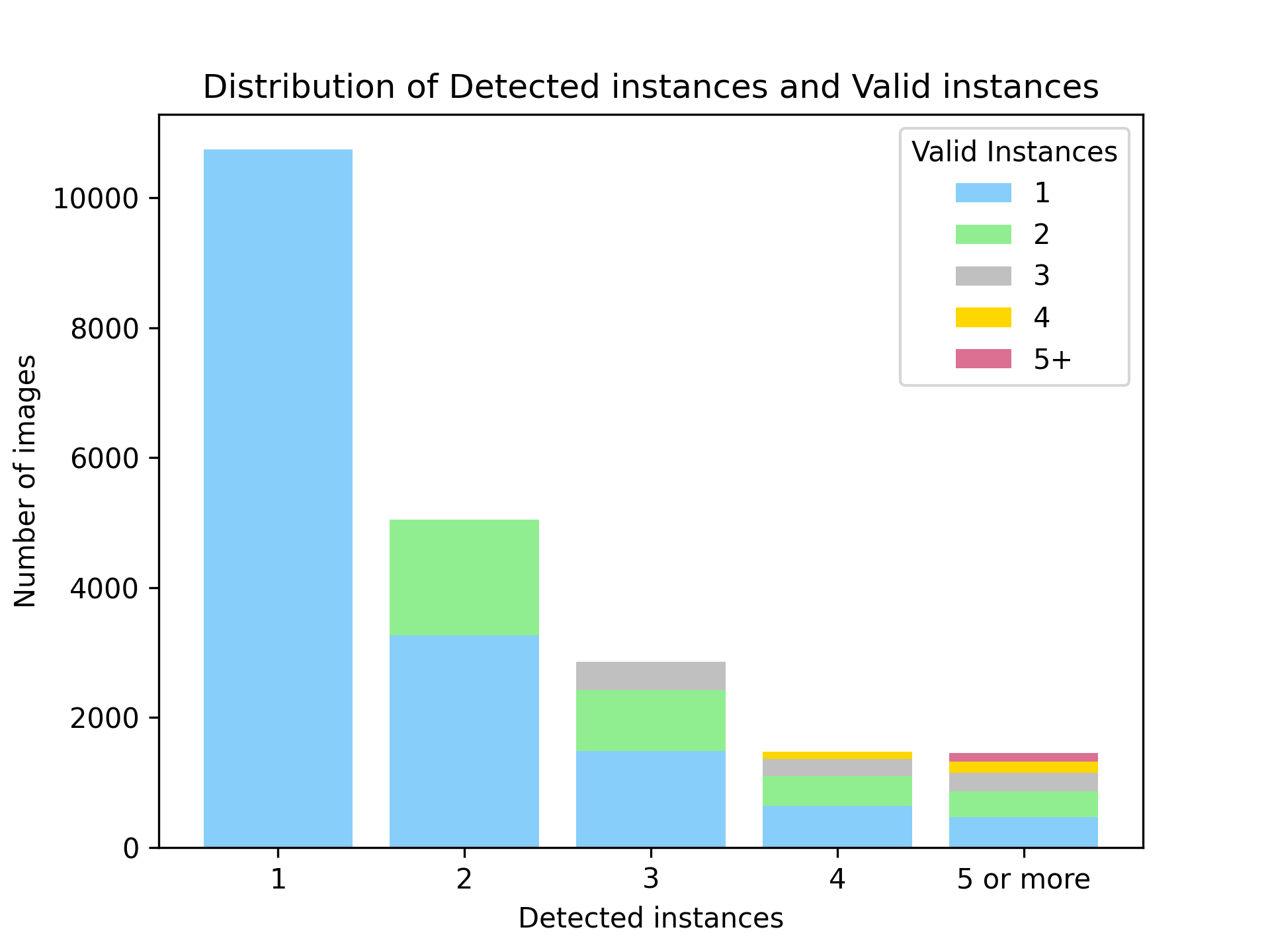}
\caption{We show the distribution of detected instances per image, as well as the proportion of valid instances.}
\label{fig:histograms}
\end{figure}

Among all the images, there are $10,752$ images with $1$ detected instance, and clearly they are all valid instances. There are $5,046$ images with $2$ detected instances, in which $1,778$ images have $2$ valid instances and the remaining $3,268$ images only have one valid instance. There are $2,853$ images with $3$ detected instances, in which $437$ images have $3$ valid instances, $938$ images have $2$ valid pairs, and $1,478$ images only have one valid instance. There are $1,476$ images with $4$ detected instances, in which $116$ images have $4$ valid instances, $268$ images have $3$ valid instances, $458$ images have $2$ valid instances, and $634$ images only have one valid instance. There are $1,448$ images with $5$ or more detected instances, in which $131$ images have $5$ or more valid instances, $171$ images have $4$ valid instances, $289$ images have $3$ valid instances, $388$ images have $2$ valid instances, and $469$ images only have one valid instance. The above information is summarized in Figure~\ref{fig:histograms}. 


\section{Conclusion}
In this work, we have constructed a large-scale real-world shadow generation dataset DESOBAv2 to supplement the existing small-scale dataset DESOBA. We hope that this dataset could contribute to shadow generation for image composition.

{\small
\bibliographystyle{ieee_fullname}
\bibliography{main.bbl}
}

\end{document}